\documentclass{article}

\usepackage{microtype}
\usepackage{graphicx}
\usepackage{subcaption}
\usepackage{booktabs} % for professional tables

\usepackage{hyperref}

\usepackage{url}            % simple URL typesetting
\usepackage{booktabs}       % professional-quality tables
\usepackage{amsfonts}       % blackboard math symbols
\usepackage{nicefrac}
\usepackage{wrapfig}% compact symbols for 1/2, etc.
\usepackage{microtype}      % microtypography
\usepackage{amsmath, amsthm} % tr`es bon mode math?matique
\usepackage{amsfonts,amssymb}
\usepackage{algorithm,algorithmic}
\usepackage{nicefrac}       % compact symbols for 1/2, etc.
\usepackage{microtype}      % microtypography

\usepackage{xcolor}
\usepackage{graphicx}
\usepackage{wrapfig,lipsum}
\usepackage{thmtools, thm-restate}
\usepackage{mathrsfs} %pour mathscr

\usepackage[shortlabels]{enumitem}
\usepackage[textsize=tiny]{todonotes}

\usepackage{bbm}

\usepackage[accepted]{icml2021}

\icmltitlerunning{17th June 2021 ICML workshop: Interpretable Machine Learning in Healthcare}

\newcommand{\A}{\mathcal{A}}
\newcommand{\bpr}{\begin{proof}}
\newcommand{\epr}{\end{proof}}
\newcommand{\be}{\begin{equation}}
\newcommand{\ee}{\end{equation}}

\newcommand{\bd}{\begin{definition}}
\newcommand{\ed}{\end{definition}}

\newcommand{\bi}{\begin{itemize}}
\newcommand{\ei}{\end{itemize}}

\newcommand{\ba}{\begin{ass}}
\newcommand{\ea}{\end{ass}}

\newcommand{\br}{\begin{remark}}
\newcommand{\er}{\end{remark}}

\newcommand{\bp}{\begin{proposition}}
\newcommand{\ep}{\end{proposition}}

\newcommand{\blm}{\begin{lemma}}
\newcommand{\elm}{\end{lemma}}

\newcommand{\bt}{\begin{theorem}}
\newcommand{\et}{\end{theorem}}

\newcommand{\bcor}{\begin{corollary}}
\newcommand{\ecor}{\end{corollary}}

\newcommand{\bex}{\begin{example}}
\newcommand{\eex}{\end{example}}

\usepackage[capitalize]{cleveref}

\crefname{assumption}{Assumption}{Assumptions}
\crefname{equation}{Eq.}{Eqs.}
\crefname{figure}{Fig.}{Fig.}
\crefname{table}{Table}{Tables}
\crefname{section}{Sec.}{Sec.}
\crefname{theorem}{Thm.}{Thm.}
\crefname{lemma}{Lemma}{Lemmas}
\crefname{corollary}{Cor.}{Cor.}
\crefname{example}{Example}{Examples}
\crefname{remark}{Remark}{Remarks}
\crefname{algorithm}{Alg.}{Algorightms}

\crefname{appendix}{Appendix}{Appendices}
\makeatletter
\def\@endtheorem{\endtrivlist}% NEW
\makeatother

\declaretheorem[name=Theorem,refname=Theorem]{theorem}
\declaretheorem[name=Lemma,sibling=theorem]{lemma}

\declaretheorem[name=Proposition,refname=Proposition,sibling=theorem]{proposition}
\declaretheorem[name=Remark,sibling=theorem]{remark}
\declaretheorem[name=Corollary,refname=Corollary,sibling=theorem]{corollary}
\declaretheorem[name=Definition,refname=Definition]{definition}

\declaretheorem[name=Example]{example}

\newcommand{\R}{\ensuremath \mathbb{R}}

\renewcommand{\P}{{\mathcal{P}}}

\let\mathsf\relax    % just to avoid a message in the log file
\DeclareRobustCommand{\mathsf}[1]{\text{\normalfont\sffamily#1}}

\newcommand{\X}{{\mathcal X}}

\newcommand{\Z}{{\mathcal Z}}

\renewcommand{\paragraph}[1]{{\bfseries #1.}}

\crefname{assumption}{Assumption}{Assumptions}
\crefname{equation}{}{}
\Crefname{equation}{Eq.}{Eqs.}
\crefname{figure}{Fig.}{Fig.}
\crefname{table}{Table}{Tables}
\crefname{section}{Sec.}{Sec.}
\crefname{theorem}{Thm.}{Thm.}
\crefname{proposition}{Prop.}{Prop.}
\crefname{fact}{Fact}{Facts}
\crefname{lemma}{Lemma}{Lemmas}
\crefname{corollary}{Cor.}{Cor.}
\crefname{example}{Example}{Examples}
\crefname{remark}{Remark}{Remarks}
\crefname{algorithm}{Alg.}{Algorithms}

\begin{document}

\twocolumn[
\icmltitle{Enabling risk-aware Reinforcement Learning for medical interventions through uncertainty decomposition}
%Uncertainty decomposition for off-policy RL and risk-aware decision-support systems



\icmlsetsymbol{equal}{*}

\begin{icmlauthorlist}
\icmlauthor{Paul Festor}{equal,ab,gh}
\icmlauthor{Giulia Luise}{equal,ab}
\icmlauthor{Matthieu Komorowski, MD}{ef}
\icmlauthor{A. Aldo Faisal}{ab,gh,cd}
\end{icmlauthorlist}

\icmlaffiliation{ab}{Brain\& Behaviour Lab: Dept. of Computing, Imperial College London}
\icmlaffiliation{gh}{UKRI Centre in AI for Healthcare, Imperial College London}
\icmlaffiliation{cd}{Brain\& Behaviour Lab: Dept. of Bioengineering, Imperial College London}
\icmlaffiliation{ef}{Dept. of Surgery \& Cancer, Imperial College London}

\icmlcorrespondingauthor{A. Aldo Faisal}{a.faisal@imperial.ac.uk}

\icmlkeywords{Machine Learning, ICML}

\vskip 0.3in
]



\printAffiliationsAndNotice{\icmlEqualContribution} % otherwise use the standard text.

\begin{abstract}
% TECHNICAL CHALLENGE.
% \textcolor{red}{placeholder}
% \aldo{this is the wrong start, spesis is only a small exmaple of where off-policy learned RL decision making systems cna come into play, just mention it as an applicaiton exmaple in the last sentnce of the abstract}

Reinforcement Learning (RL) is emerging as tool for tackling complex control and decision-making problems. However, in high-risk environments such as healthcare, manufacturing, automotive or aerospace, it is often challenging to bridge the gap between an apparently optimal policy learned by an agent and its real-world deployment, due to the uncertainties and risk associated with it. Broadly speaking RL agents face two kinds of uncertainty, 1. aleatoric uncertainty, which reflects randomness or noise in the dynamics of the world, and 2. epistemic uncertainty, which reflects the bounded knowledge of the agent due to model limitations and finite amount of information/data the agent has acquired about the world. These two types of uncertainty carry fundamentally different implications for the evaluation of performance and the level of risk or trust. Yet these aleatoric and epistemic uncertainties are generally confounded as standard and even distributional RL is agnostic to this difference. Here we propose how a distributional approach (UA-DQN) can be recast to render uncertainties by decomposing the net effects of each uncertainty . We demonstrate the operation of this method in grid world examples to build intuition and then show a proof of concept application for an RL agent operating as a clinical decision support system in critical care.
 
% Sepsis is a life-threatening infection that proves fatal in almost a third of patients. Treating sepsis is clinically very challenging and there is no well-defined and confirmed treatment. Treatment of sepsis can be interpreted as a sequential decision making problem and - recently - reinforcement learning methods applied on retrospective data have shown promising results that could eventually identify individualized and clinically effective treatment decisions. However, the road towards the actual deployment of these methods in hospitals is still long. A detailed analysis of the uncertainties at play is key to understand not only the potential but also the limitations of computed policies and to unlock the first steps towards real-life deployment.

% We propose to use a combination of Bayesian inference and distributional Reinforcement Learning to incorporate different sources of uncertainties. We show preliminary results on MIMICIII dataset and outline future directions.
% RL for recommender systems enabled us to learn them for ubiquitous EHR data. But what about uncertainy and risk and their deployment.

% have been studied in order to identify individualized and clinically interpretable treatment decisions that could improve patient outcomes.
\end{abstract}

\section{Introduction}
The increasing availability of electronic healthcare records containing both patient state and clinician's treatment actions opened up the opportunity to machine learn  the decision making policies for patient treatments. Therefore, off-policy Reinforcement Learning {RL} for optimal sequential clinical decisions have become the paragon for AI-based interventions \cite{komorowski2018artificial, gottesman2019guidelines}. In general RL shows very promising performance and potential to exceed human expert performance when tested on retrospective or simulated data \cite{ernst2006clinical,bothe2013use,lowery2013towards,li2020optimizing,liu2020reinforcement}.
First reinforcement learning-derived systems that autonomously intervene on patients have already been deployed successfully in rehabilitation settings \cite{wannawas2021robot} where risks are controlled, but the hospital-based use-case of RL driven intervention at the bed side involves greater risks, due to health state of patients and potential negative impact of any intervention. There is therefore now a growing need to direct development of RL systems for healthcare that are risk and safety aware \emph{a priori} by design. This requires RL algorithms to be aware of noise and uncertainty in their operation, and to make these critical performance interrogatable, so as to ultimately allow developers, regulators and users to specify safety and trust margins in their operation.

One of the areas where RL has shown great potential is the treatment of sepsis in critical care, such as the management of sepsis, which kills almost a third of patients \cite{stevenson2014two} and yet optimal  decision making policies are unclear. RL methods applied to retrospective data have shown promising results that could theoretically identify individualized and clinically effective treatment decisions that significantly reduce mortality of patients \cite{komorowski2018artificial}. Here, the RL agent's actions, referred to cheekily as AI Clinician, are communicated as recommendations (clinical decision support system, CDSS)
to an expert, e.g. a critical care clinician, who may then choose to implement or ignore. Interestingly this intensive care application has rapidly evolved to become a popular use case for  reinforcement learning in healthcare, in parts due to the public availability of high density, high quality data sets, as well as the immediate usefulness of reinforcement learning over that of supervised learning. Consequently, clinical trials of RL systems are now being ramped up to test such systems in hospitals, there are limitations on the interpretability, safety and trust of that have need to be overcome before general deployment can be considered. 

Broadly speaking, there are two main sources of uncertainty that limit the capability of an RL agent's optimal policy: First, noise or randomness in the dynamics of the world and the agent. This uncertainty is measurable by repeating the same action and observing variability from trial to trial \cite{faisal2008noise}. This so called aleatoric uncertainty (from the Latin \emph{alea} for "dice") captures thus randomness in the system. In the healthcare case this can be physiological processes, the effect of treatments or drugs, or the outcomes of healing. These are factors that the agent can mitigate by working around them, but ultimately has no direct control over. In healthcare treatments there may be many settings where an action or intervention may have a probabilistic outcome and it is important for a human stake holder to understand if a proposed outcome's success is affected by chance.

Second, we have epistemic uncertainty (from the Greek word \emph{episteme} for knowledge) that captures how well the model has learnt from finite data to solve the problem and how well the model can fit the problem. In the healthcare case this could reflect that the algorithm is aware that it has limited data  of state space regions or state transitions e.g. when facing a new disease or when a drug has not been often prescribed to a particular type of patient. Epistemic uncertainty is essential for recommender systems safety, as this allows human stake holders to weight an AI systems recommendation against human judgement or prompt seeking expert human opinions.

These two types of uncertainties are typically not accessible in conventional 
RL models. More advanced probabilistic RL models which use posterior 
distributions over learned parameters in Bayesian RL \cite{ghavamzadeh2016bayesian} or in distributional reinforcement learning which models distributions over 
the returns \cite{bellemare2017distributional} also confound or conflate the two types of variability. 
However, being able to decompose these two components would have immediate 
benefits for the interpretation of RL agent treatment recommendations:
For example, high aleatoric and low epistemic uncertainty in a given 
state would mean that the model is aware that this state is highly stochastic, so experts should be particularly careful about the patient's evolution here, 
whereas high epistemic uncertainty would mean that the model has not 
seen enough data or has not been able to converge well 
enough in that state to give an informed recommendation, so clinicians should 
not consider the model output as very 
valuable.

% % 
In the following we show how using existing concepts and framework we can build RL agents that decompose the uncertainties associated with their state-action representations. We illustrate the behaviour of our model on toy grid world setups, and show what both uncertainties represent through intuitive computer experiments. We  port this approach and apply it in an RL clinical decision support system.

\section{Background}

Let $\X$ and $\A$ be state and actions spaces, respectively and let $\mathcal{Z} := \X\times \A$ denote the product space. We consider an MDP framework where we have transition kernels $p:\X \times \A \rightarrow \P(\R \times \X)$. The full MDP is given by a collection $(X_t,A_t,R_t)_{t \geq 0})$, where $(X_t)_{t\geq0}$ is the sequence of states taken from the environment, $(A_t)_{t\geq0}$ is the sequence of actions taken by the agent and $(R_t)_{t\geq0}$ the sequence of rewards. We will consider either deterministic policies $\pi:\X\rightarrow \A$ or stochastic policies $\pi:\X\rightarrow \P(\A)$. The \textit{return} of a policy $\pi$, starting at an initial state $x\in\X$ and initially taking an action $a\in \A$ is the random variable given by the sum of discounted rewards, $\sum_{t=0}^{\infty} \gamma^t R_t \Big\vert \,\, X_0 = x,\,\, A_0= a$.
Additionally, the distribution of the return of policy $\pi$ and initial state-action
pair $(x, a) \in \Z$  is \cite{rowland2018analysis} 
\begin{equation}\label{eq:return_distribution}
    \eta_\pi^{(x,a)} = \textnormal{Law}_\pi\Big[  \sum_{t=0}^{\infty} \gamma^t R_t \Big\vert \,\, X_0 = x,\,\, A_0= a\Big].
\end{equation}

% Off-policy RL methods produce policies which are learnt on retrospective data and commonly evaluated with off-policy policy evaluation techniques \cite{thomas2015high_OPE, hanna2016bootstrapping_OPE}. These evaluations give estimates of a new policy's performance using transitions data produced by an original policy, usually the clinician's behaviour policy in an hospital setup.  The actual performance of an agent can only truly be assessed when deployed at the bedside, which is more risk-involved, and proper uncertainty qualification and measures can help bridging this gap.

 Having access to the distribution of returns rather than just the expectation as in standard RL leads to additional information that plays a crucial role when quantifying \textit{uncertainties}. Here, distributional RL aims to learn the distribution of returns $\eta_\pi^{(x,a)}$ associated with taking action $a$ in state $x$ and then following a policy $\pi$. The idea is to consider the whole distribution of the return rather than the expectation  \cite{morimura2010parametric,morimura2010nonparametric}, where the return density is estimated in order to handle a variety of risk-sensitive and risk-averse criteria. This approach reached breakthrough performance in Atari Games benchmarks \cite{bellemare2017distributional}.
% \aldo{LINK?}
In terms of algorithmic developments, in the past few years, several distributional RL approaches have been developed to learn the return distributions $\eta^\pi(s,a)$, as \citet{bellemare2017distributional, dabney2018distributional, yang2019fully} to name a few. In \cite{dabney2018distributional}, the return distribution $\eta^{\pi}(x,a)$ is approximated  by $N$ quantiles $\tau_i = i/(N+1)$, with $i\in[1,N]$ with corresponding values $\boldsymbol{q} = (q_1,\dots,q_N)$. The values of the quantiles can be learned leveraging the Huber loss and the Bellman target as in the QR-DQN algorithm by Dabney et al. \cite{dabney2018distributional}.
We adapt here the approach developed  by \cite{clements2019estimating} to combine Bayesian inference and distributional RL to then incorporate during training and then render the decomposition of the different sources of uncertainty. The epistemic and aleatoric uncertainties are incorporated as additional information during the training procedure in a variant of Quantile-Regression DQN. 
In order to decouple aleatoric and epistemic uncertainties one can learn the quantiles of the return distribution in a Bayesian fashion; to learn the value of a given quantile $\tau$ of $\eta^{\pi}(x,a)$, a neural network is used with parameters $\boldsymbol{\theta}$ that return a value $y(x,a,\boldsymbol{\theta})$. Given a set of samples $\mathcal{S}:=(s_1, \dots, s_k)$ from $\eta^{\pi}(x,a)$ (or, in practice, from the corresponding Bellman target), the likelihood of $\mathcal{S}$ given $\mathsf{\theta}$ is defined as:
\begin{equation}\label{eq:likelihood}
    P(\mathcal{S}\mid \boldsymbol{\theta}) = \sum_{j=1}^k\sum_{i=1}^N \ell_{\tau_i}(s_i - y_i(\boldsymbol{\theta}, x,a))
\end{equation} 
where $\ell_{\tau_i}$ is based on the asymmetric Laplace distribution - a standard approach in Bayesian Quantile Regression \cite{yu2001bayesian}. \citet{clements2019estimating} proceed to show how obtaining uncertainty estimates can help boost an agent's training process. Using this line of thought, the following estimators for the epistemic and aleatoric uncertainties can be obtained (here using variance as proxy for uncertainty):
\begin{equation*}
    \sigma^2_{\textnormal{epistemic}} := \mathbb{E}_{i\sim U\{1,N\}}[ \textnormal{var}_{\boldsymbol{\theta}\sim P(\boldsymbol{\theta}\mid \mathcal{S})}(y_i(\boldsymbol{\theta}, x,a))]\\
\end{equation*}
This captures the variability in the return distribution estimate due to model learning by averaging over trajectories and taking the variance of the quantile estimates with respect to the posterior on policy parameters.
\begin{equation*}
\sigma^2_{\textnormal{aleatoric}} := \textnormal{var}_{i\sim U\{1,N\}} [\mathbb{E}_{\boldsymbol{\theta}\sim P(\boldsymbol{\theta}\mid \mathcal{S})}(y_i(\boldsymbol{\theta}, x,a))]
\end{equation*}
This captures the variability in the return distribution estimate due to stochasticity in the environment by averaging over posterior policy parameters and taking the variance of the quantile estimates with respect to trajectories.

We argue and visualise here how this approach can be used for uncertainty decomposition to render safety-conscious and risk-aware recommender systems -- which frame and  put into a risk-conscious context the recommendations of a CDSS.
% \subsection{Decoupling of uncertainty: theory and real-world implications}

We argue that both epistemic and aleatoric uncertainty are of crucial importance in medical applications and that an accurate estimate of the two uncertainties is fundamental in order to unlock the practical deployment of reinforcement learning in healthcare. Decoupling them would have profound implications for the interpretability and accountability of the output of a CDSS. Equally, the trust that users will in  CDSSs will only increase if they can provide some degree of confidence in their output or manage expectations of outcomes, e.g. when the aleatoric uncertainty is high. Stake holders will be interested to know that a recommended action comes with a low confidence of success because of rare patients characteristics (leading to high epistemic uncertainty) or that the uncertainty about the future evolution of a given patient is high (high aleatoric uncertainty).
Another positive effect of quantifying uncertainty will be to enable clinicians to focus their efforts on patients where uncertainty is the highest, among a given cohort of patients in the intensive care unit, leaving those with high confidence "under the care" of the AI.

\section{Results}
In this section we present experimental results to support our claims that modelling and disentangling both sources of uncertainties in clinical setting is potentially very informative and crucial for clinical deployments.

\subsection{Toy experiments: grid worlds}

We first introduce two grid worlds to validate this approach of computing proxies for aleatoric and epistemic uncertainty estimates.

\begin{figure}[htb]
\centering
\includegraphics[width=0.35\textwidth]{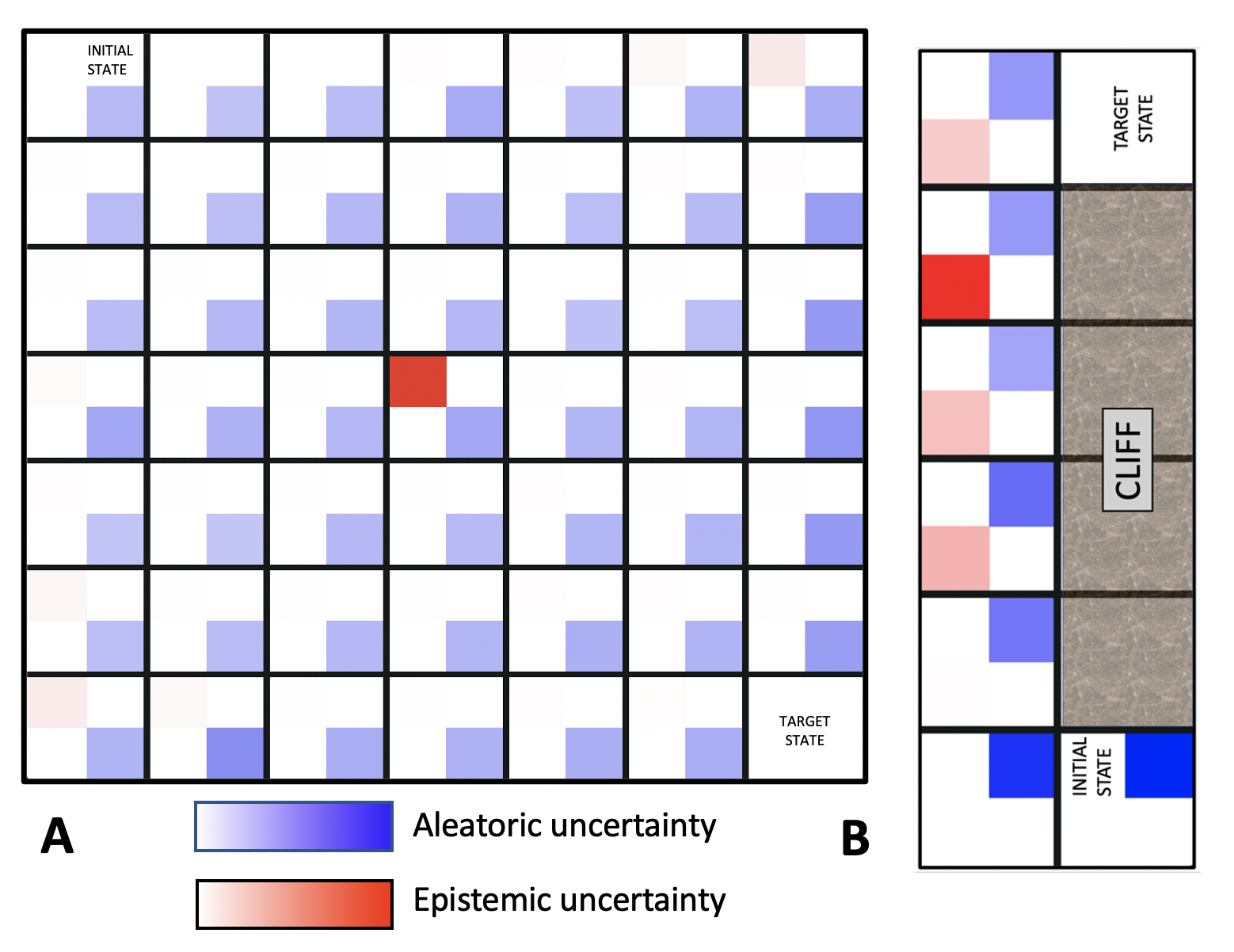}
%\aldo{update this figure as per caption}
\caption{(A) Representation of the agent's uncertainty estimates for a $7 \times 7$ grid world. The dark lines are separating the grid world states. The agent is given positive reward for successfully going from the initial state to the terminal state. In each state's square the top red color box represents the epistemic uncertainty and blue box at the bottom right represents the aleatoric uncertainty (more vivid colour corresponding to more uncertainty normalised across all states). To aide intuition we scaled the value of aleatoric uncertainty here with the computed values for the same grid world where the state transition probabilities are uniform random for any action (random walker agent). (B) 
Representation of the agent's uncertainty estimates in a $2\times 6$ cliff-walking high-risk grid world where states are subject to wind which pushed the agent into the cliff with $20\%$ probability.}
\label{fig:gridworld_uncertainties}
\end{figure}

First, epistemic uncertainty should come from faults in model learning: it corresponds to the part of the uncertainty due to lack of data or poor model convergence. To build intuition of our approach we visualise here a  $7 \times 7$ deterministic grid world Fig.~\ref{fig:gridworld_uncertainties}.A. 

Epistemic uncertainty is high for states where we have observed less data. Our approach should flag regions of high epistemic uncertainty. 
To validate this we ran computer experiments where observations of transition from the central state -- in grid coordinates (4,4) -- were artificially reduced (by including these transition with only a $1\%$ chance into the replay buffer). Correspondingly our approach flagged that central state as one of high epistemic uncertainty (bright red box in central state, Fig.~\ref{fig:gridworld_uncertainties}.a).
In contrast the aleatoric uncertainty is uniformly distributed and small , reflecting the 
The differences in aleatoric uncertainty in this figure are only noise, there is no stochasticity in this world, the values are very close but normalisation emphasises differences in the visualisation.

We proceeded in a similar way to validate the behaviour of UADQN's aleatoric uncertainty estimate. Aleatoric uncertainty should be a measure of how much the variability in the reward distribution is due to intrinsic stochasticity of the world. We trained the agent to act in  a $2\times 6$ grid world where the start state (Fig.~\ref{fig:gridworld_uncertainties}.b), at the left of the bottom row, is separated from the goal state by a cliff. When walking along the cliff edge, wind will push the agent down wards with $20\%$ probability (irrespective of the action chosen). The agent gets positive reward for reaching the goal, and negative reward for falling off the cliff. 
 
This wind mechanic introduces stochasticity in the world dynamics and should impact the aleatoric uncertainty values. Figure \ref{fig:gridworld_uncertainties}.b visualises the uncertainty estimates in the cliff grid world, and confirms that higher aleatoric uncertainty is observed in states where stochasticity in the environment will have the most impact on the return distribution. The closer to the goal, the less chances the wind has to push the agent into the cliff, thus the less aleatoric uncertainty. This figure also supports the good behaviour of the epistemic uncertainty estimates since higher epistemic uncertainty is observed on the top-right states, which were visited less often by the agent during training due to falling off the cliff.

\subsection{Intensive care based clinical recommender system}

Next, we applied the uncertainty aware agent to an MDP adapted from the original MIMIC-III critical care dataset and AI clinician work by \cite{komorowski2018artificial}. We are here interested in the challenge of learning optimal treatment strategies for septic patients from retrospective data, corresponding to the dosing of vasopressor drugs and IV fluids for successive 4-hour time windows. The aim is to render the estimates of aleatoric and epistemic uncertainty for the drug dosage  as part of the treatment recommendations.
%In contrast to The MDP uses both intermediate rewards based on organ failure (difference of organ failure score between subsequent time points) and terminal rewards (positive for patient survival, negative for patient death). Given that not all actions were available to the agent in all states, optimal actions were selected by maximising over the $Q$-values of available actions.
After convergence, the agent provides estimates of aleatoric and epistemic uncertainty for each state action pair. 
The state-action space is sparse and vast (752 states $\times$ 25 actions). We observe variability in the two types of uncertainties across the state space that are distinct (Fig.~2.a). Consistent with intuition, plotting the epistemic uncertainty is clearly anticorrelated with state visitations in the retrospective data (Fig.~2.b).
States visited the most often are mostly associated with low epistemic uncertainty and high epistemic uncertainty only appears in less visited states. These results are in line with the behaviours observed in the grid world examples, supporting out approaches use case further.

\begin{figure}[tbp]
\centering
\includegraphics[width=0.45\textwidth]{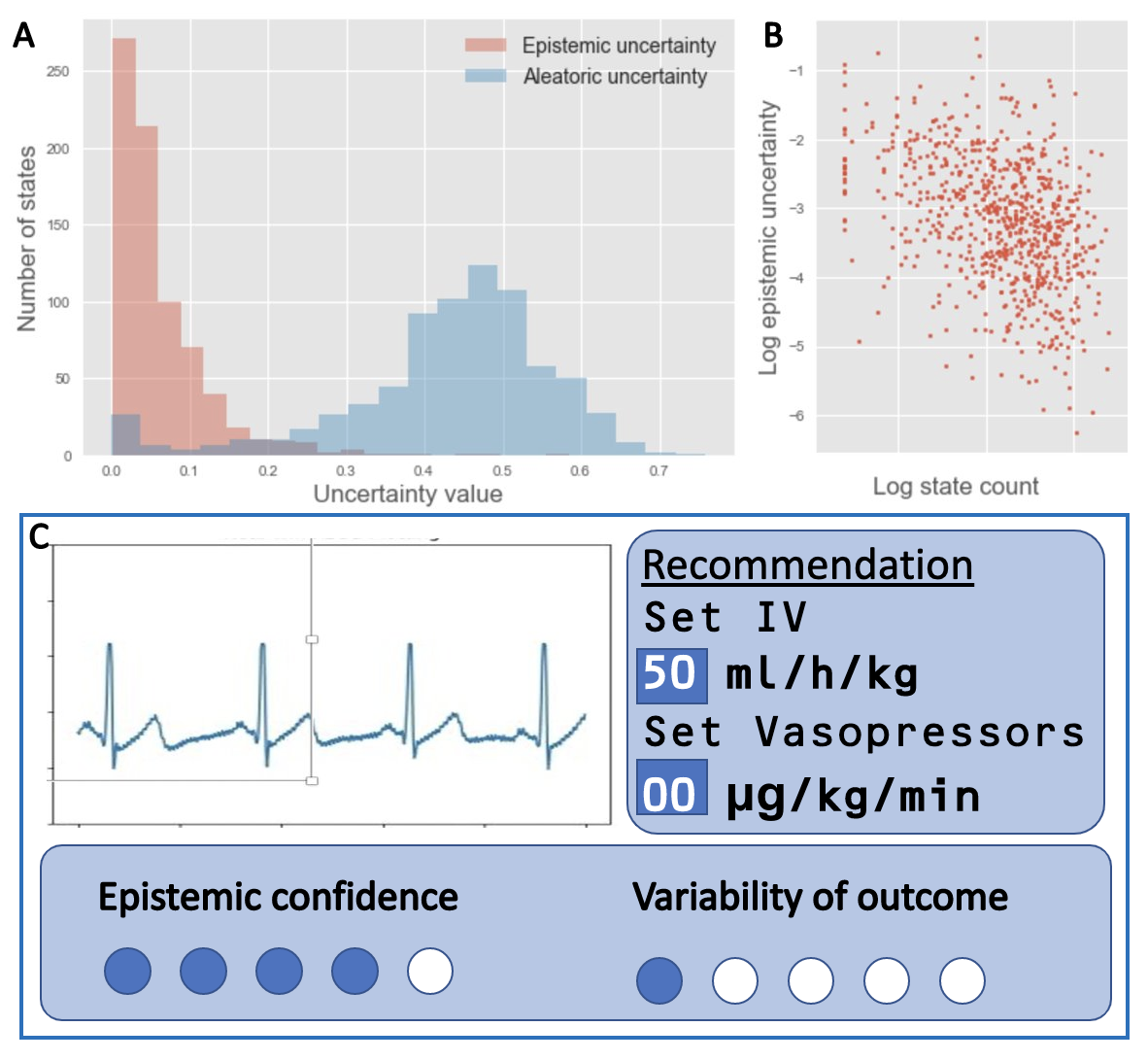}
\caption{(A) Histogram representing the distribution of the agent's uncertainty estimates on the intensive care data based MDP. (B) Scatter plot of the estimated epistemic uncertainty against state visitation, obtained by sampling trajectories of the learnt policy, in a log-log scale. (C) Sketch of an interface for the system to be seen by the clinical team. Model recommendations would be nuanced by a confidence score (derived from epistemic uncertainty) and the system would evaluate variability in the patient outcome (derived from aleatoric uncertainty).}
\label{fig:MIMIC_results}
\end{figure}

\section{Discussion and Conclusion}

The MDP for intensive care used in this work is a simplification of the real environment in which the state space is continuous and patient dynamics are more complex, so there still is a challenge in bringing the estimation of uncertainties forward into these complex settings.
This is just the beginning of investigating the features of the data and the derived uncertainties in RL agents. The aim of this paper is to raise awareness around the benefits that disentangling uncertainties can have on RL-based decision support systems and show a proof of concept for such model.

This work presents an uncertainty and risk-aware approach in the deployment of RL for clinical applications. We think that exposing the sources of epistemic and aleatoric uncertainties when communicating recommendations to the end-users and stake holders is crucial. High-quality explainability in reinforcement learning are hard to design. Previous approaches to RL explanations focused on rendering a causal explanations why a specific policy was learned (e.g. \cite{madumal2020explainable})  or by breaking down an agent's policy into human-understandable intermediate policies
%that explain how a overarching goal is achieved 
\cite{beyret2019dot}. Here we focused on interpretabiltiy of a recommendation in terms of the uncertainties involved. 
% We believe this information will be important in giving human users agency on how to act on an RL recommender systems' output.
We believe this information will be important for human users to interact with RL recommender systems' outputs.

\section*{Acknowledgements}

PF was supported by a PhD studentship of the UKRI Centre for Doctoral Training in AI for Healthcare (EP/S023283/1).  AAF was supported by a UKRI Turing AI Fellowship (EP/V025449/1). This project is independent research funded by the National Institute for Health Research (Artificial Intelligence, [“Validation of a machine learning tool for optimal sepsis treatment.”, AI\_AWARD01869]). The views expressed in this publication are those of the author(s) and not necessarily those of the National Institute for Health Research or the Department of Health and Social Care.

\nocite{langley00}

\bibliography{example_paper}
\bibliographystyle{icml2021}

% \appendix
% \section{Do \emph{not} have an appendix here}

% \\textbf{\emph{Do not put content after the references.}}
%
% \Put anything that you might normally include after the references in a separate
% \supplementary file.

% \We recommend that you build supplementary material in a separate document.
% \If you must create one PDF and cut it up, please be careful to use a tool that
% \does not alter the margins, and that does not aggressively rewrite the PDF file.
% \pdftk usually works fine. 

% \\textbf{Please do not use Apple's preview to cut off supplementary material.} In
% \previous years it has altered margins, and created headaches at the camera-ready
% \stage. 
%%%%%%%%%%%%%%%%%%%%%%%%%%%%%%%%%%%%%%%%%%%%%%%%%%%%%%%%%%%%%%%%%%%%%%%%%%%%%%%
%%%%%%%%%%%%%%%%%%%%%%%%%%%%%%%%%%%%%%%%%%%%%%%%%%%%%%%%%%%%%%%%%%%%%%%%%%%%%%%

\end{document}